\documentclass[10pt,conference,a4paper]{IEEEtran}
\usepackage{booktabs}
\usepackage{tabu}
\usepackage{threeparttablex}

\ifCLASSOPTIONcompsoc
  \usepackage[nocompress]{cite}
\else
  \usepackage{cite}
\fi
\ifCLASSINFOpdf
  \usepackage[pdftex]{graphicx}
  \DeclareGraphicsExtensions{.pdf,.jpeg,.png}
\else
\fi
\usepackage{amsmath}

\DeclareMathOperator*{\argmax}{arg\,max}
\usepackage{array}

\ifCLASSOPTIONcompsoc
  \usepackage[caption=false,font=footnotesize,labelfont=sf,textfont=sf]{subfig}
\else
\usepackage[caption=false,font=footnotesize]{subfig}
\fi
\begin{document}
%
\title{Annotation Order Matters: \\Recurrent Image Annotator for Arbitrary Length Image Tagging}

\author{\IEEEauthorblockN{Jiren Jin}
\IEEEauthorblockA{The University of Tokyo\\
7-3-1 Hongo, Bunkyo-ku, Tokyo, Japan\\
Email: jin@nlab.ci.i.u-tokyo.ac.jp}
\and
\IEEEauthorblockN{Hideki Nakayama}
\IEEEauthorblockA{The University of Tokyo\\
7-3-1 Hongo, Bunkyo-ku, Tokyo, Japan\\
Email: nakayama@nlab.ci.i.u-tokyo.ac.jp}
}


%


\maketitle

\begin{abstract}
\boldmath Automatic image annotation has been an important research topic in facilitating large scale image management and retrieval. Existing methods focus on learning image-tag correlation or correlation between tags to improve annotation accuracy. However, most of these methods evaluate their performance using top-$k$ retrieval performance, where $k$ is fixed. Although such setting gives convenience for comparing different methods, it is not the natural way that humans annotate images. The number of annotated tags should depend on image contents. Inspired by the recent progress in machine translation and image captioning, we propose a novel Recurrent Image Annotator (RIA) model that forms image annotation task as a sequence generation problem so that RIA can natively predict the proper length of tags according to image contents. We evaluate the proposed model on various image annotation datasets. In addition to comparing our model with existing methods using the conventional top-$k$ evaluation measures, we also provide our model as a high quality baseline for the arbitrary length image tagging task. Moreover, the results of our experiments show that the order of tags in training phase has a great impact on the final annotation performance.
\end{abstract}

%
\IEEEpeerreviewmaketitle

\section{Introduction}

Image annotation is a task to associate multiple semantic tags regarding to the contents of images. With the rapid development of Internet and social web applications, the amount of online images created by users is continuously increasing. The large amount of images brings a heavy burden for image management and retrieval. Since the major approaches for people to search or to index images are through referring to the associated tags, it is a necessary step to annotate these images with proper tags. However, manually annotating images is an expensive and labor intensive work for human beings. Hence it is better if we can learn a model from available image-tag samples and use the model to automatically label new images with keywords (tags) from the annotation vocabulary. In fact, this kind of technique is called automatic image annotation (AIA)~\cite{zhang2012review}, which has been an important research topic in computer vision for decades.

\begin{figure}[!t]
\centering
\includegraphics[width=3.5in]{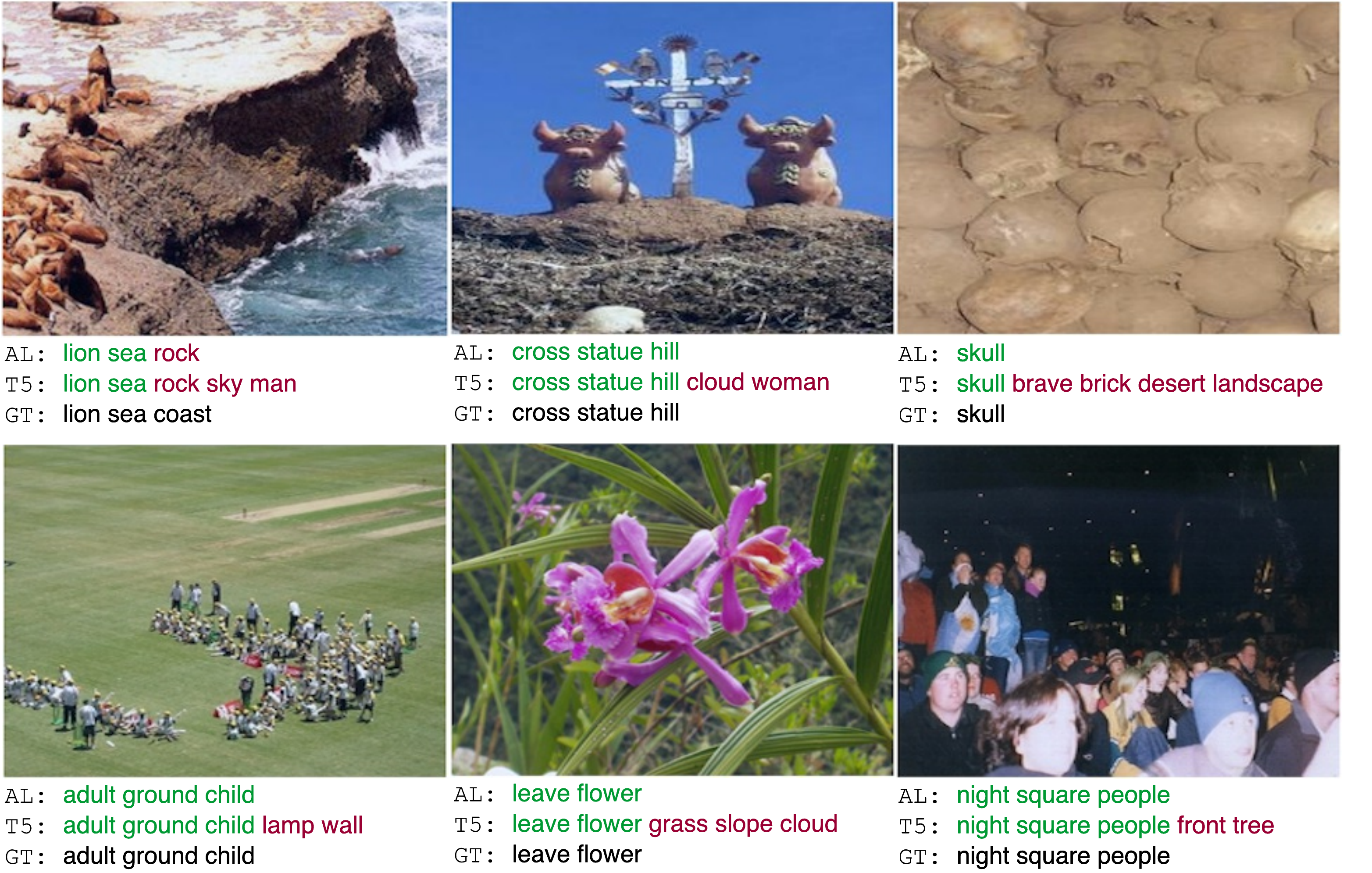}
\caption{Examples of results showing differences between Top-5 annotations and arbitrary length annotations. Top-5 annotations tend to generate more false positives (red). AL: arbitrary length, T5: Top-5, GT: ground truth.}
\label{fig_top5_arbitrary}
\end{figure}

Previous researches focus on learning the image-to-tag correlation as well as tag-to-tag correlation to improve the annotation performance. Although much progress has been made in the research community, most of the existing methods overlooked a fundamental philosophy of recognition: recognizing the right things. A common conventional evaluation setting has a fixed annotation length $k$, and a typical $k$ value $5$ has been used in many previous methods~\cite{feng2004multiple, makadia2008new, guillaumin2009tagprop, verma2012image} for the ease of comparison. However, we argue that this convention can be insufficiency in previous work, since it is not the normal way that we humans annotate images, and the assumption of fixed annotation length is not the fact of realistic images either, as shown in Figure \ref{fig_top5_arbitrary}. Therefore arbitrary length annotation is required for more reasonable annotation results. For top-$k$ predictions, traditional methods simply select the $k$ tags with highest prediction scores. For arbitrary length annotation, it is possible to easily imagine a naive extension that is to threshold the prediction scores. However, finding a good threshold is more difficult than merely setting a hyper-parameter as we might expect, because the optimal threshold can actually be dependent on each different image.

\begin{figure}[!t]
\centering
\includegraphics[width=3.5in]{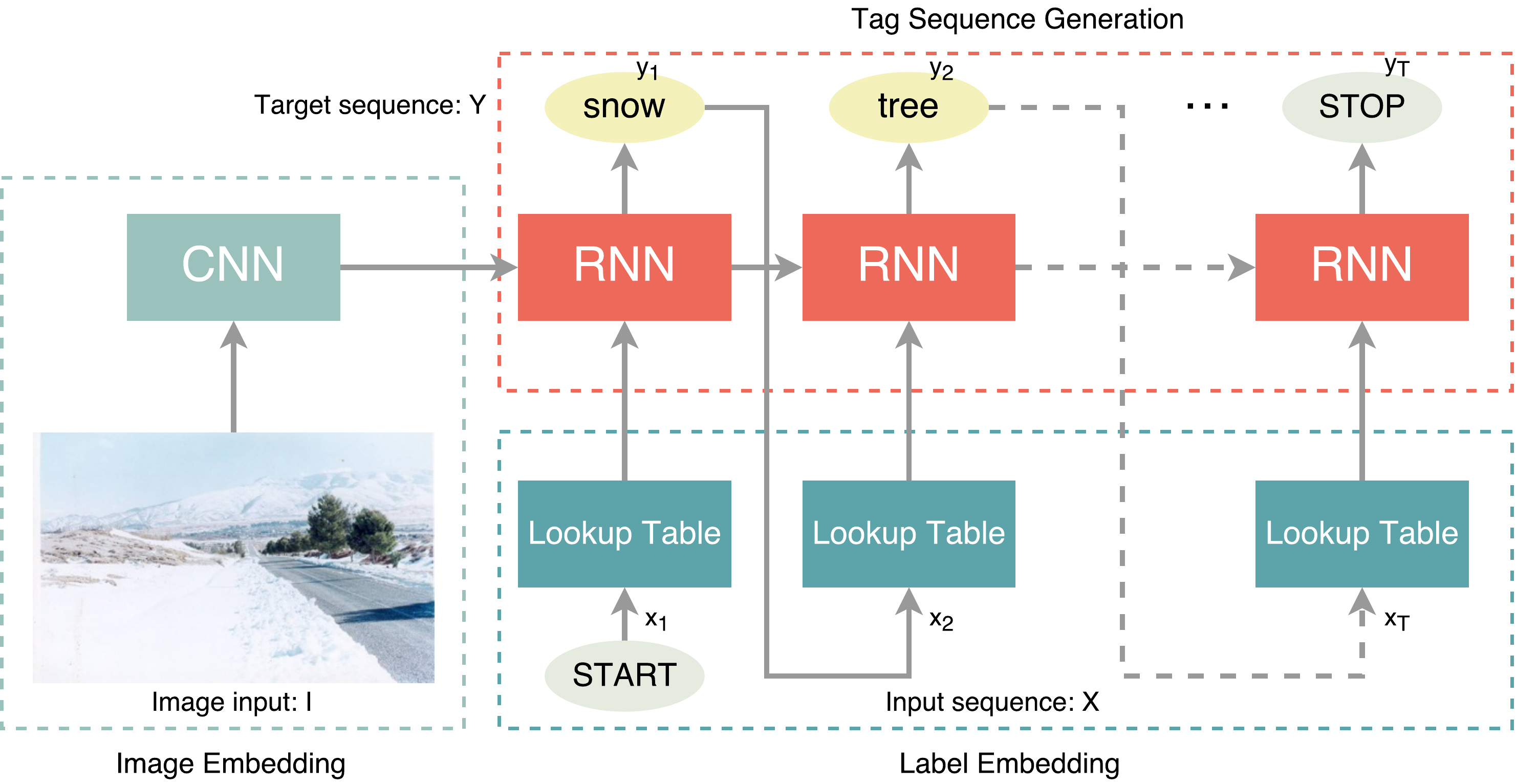}
\caption{General architecture of RIA model. In test phase, once the RIA model receives the input image $I$, and is triggered by the \textit{START} signal, it predicts the first output tag. Then it starts a loop that uses previous output as input of the next time step, predicting the tag sequence Y recursively. The loop will continue until the \textit{STOP} signal is predicted.}
\label{fig_model}
\end{figure}

Instead of struggling to find the appropriate threshold, we want to import an explicit mechanism to model the annotation length, for which we originally form the image annotation task as a sequence generation problem. Therefore we propose a novel model called Recurrent Image Annotator (RIA) that jointly uses Convolutional Neural Networks and Recurrent Neural Networks (RNN) for predicting tag sequences. In the annotation phase, we just use an image as the initial input of RIA and then it will automatically generate annotation tags one by one, as shown in Figure \ref{fig_model}. The idea is inspired by recent success of RNN in machine translation~\cite{bahdanau2014neural}, and especially in image captioning~\cite{mao2014explain, vinyals2015show}, where the task is to generate natural language sentences from images. The advantages of using RNN do not only include its nature to generate varied length outputs, but also its ability to refer to previous inputs when predicting the current time step output. Such ability allows RNN to exploit the correlations of both image-to-tag and tag-to-tag. 

Now we have a CNN to extract image visual features, and an RNN to generate the tag sequence from the visual features, what do we need next? The answer is: an order. Both machine translation and image captioning aim to generate sentences, which have a natural order available for the RNN model to learn from. Unfortunately, in our image annotation task, there is no natural order available. Instead, we have to choose or learn an order to make our proposed model actually work. 

Just like sentences obey the language rules to form the order, we believe that there exist intrinsic ``language rules" for tags to form an order to describe an image. There are two points for an order to be good in our task. First, the order ``rule" should be based on semantic image and tag information. Second, tag sequences in each training example should follow the same rule to be sorted, since only in this way can the model learn the ``rule" from the training examples, and further generalize the prediction on the test images.

To facilitate the training of our model as well as testing the importance of tag orders, we propose several strategies to provide tag orders. And we compare the performance of our model with different tag orders in the experiments.

The main contributions of our work are as follows:
\begin{enumerate}
\item To our best knowledge, our work is one of the first\footnote{We found~\cite{wang2016cnnrnn} became publicly available on arXiv.org after we finished our work. Though there are several similar ideas existing in both papers, the focuses and motivations of ours are different. We pay more attention to the annotation length, and the tag sequence order used in training phase.} to form image annotation task as a sequence generation problem, and we propose a novel RNN based model Recurrent Image Annotator to handle image annotation work. 
\item We analyze the insufficiency in existing methods that they do not pay enough attention to generate image dependent number of tags, which should be a natural requirement in realistic tasks. We propose our RIA model as a high quality baseline for comparing the performance on arbitrary length image tagging. We hope that our work can help and encourage future work on this new task. 
\item We propose and evaluate several orders for sorting the tag inputs of RIA model, and show the importance of tag order in the tag sequence generation problem.
\end{enumerate}


\section{Related Work}
In this section, we review previous work in AIA and introduce previous work related to our RIA model, i.e., CNN and RNN.

\subsection{Automatic Image Annotation}
Generally the existing methods of RIA can be grouped into three categories: generative models, discriminative models, and nearest neighbor type models. Generative models minimize the generative data likelihood based on topic models~\cite{duygulu2002object}, where each topic is a distribution over image features and annotation tags, or mixture models~\cite{carneiro2007supervised, feng2004multiple, lavrenko2003model}, where the models define a joint distribution over image features and annotation tags. Different from generative models, discriminative models~\cite{cusano2003image, grangier2008discriminative} focus on directly learning a classifier for tag prediction, and recently CNN based multi-label classification models have been proposed~\cite{gong2013deep, wei2014cnn}. Another simple but powerful group of models are k-nearest-neighbor (KNN) based models~\cite{makadia2008new, guillaumin2009tagprop, verma2012image}, which also benefit from metric learning of multiple hand-crafted visual features.

\subsection{Convolutional Neural Networks}
The first step in AIA is to extract effective and efficient visual features from raw image pixels. Traditional methods usually use hand-crafted global or region based image features, or the combination of them~\cite{guillaumin2009tagprop}, while recent researches indicate that features extracted from Convolutional Neural Networks (CNN)~\cite{lecun1998gradient, krizhevsky2012imagenet} have significantly superior performance over these hand-crafted features on single-label image classification task~\cite{simonyan2014very, he2015deep}. However, the recent work~\cite{murthy2015automatic} show that deep CNN features do not outperform handcrafted features a lot in the traditional methods. We think one of the possible reasons is that the benefit from metric learning on multiple hand-crafted features is lost. Another problem is that currently there is no suitable loss function that can handle multi-label image classification perfectly for CNN models (for single-label classification task defining the optimal loss is trivial). 


\subsection{Recurrent Neural Networks}
Recurrent Neural Networks (RNN) are networks with loops, which can be treated as multiple copies of the same network that are connected by passing messages (state) to the successor. However, the original architecture of RNN is difficult to train for long sequences due to gradient exploding and vanishing problem~\cite{pascanu2012difficulty}. The gradient exploding problem can be easily coped with gradient clipping, i.e., limiting the absolute value of gradients. The vanishing problem is more difficult to handle, therefore several variants of RNN have been proposed for solving the problem of long term dependencies, for example, LSTM~\cite{hochreiter1997long} and GRU~\cite{cho2014learning}. These RNN variants use hidden cell states and gate functions to control how information from each previous time step is combined and propagated, and have been proved to work better than vanilla RNN~\cite{chung2014empirical}. 

We choose LSTM as our RNN sub-module just because it has been widely used and tested. Recent researches~\cite{chung2014empirical} compare LSTM and GRU, showing that they have similar performance. In our RIA model, RNN is used as a decoder to decode tag sequence from image input, and is the crucial part to predict arbitrary length of annotation results.

\section{Recurrent Image Annotator}
In this section, we describe the entire model architecture first, and then explain the details of each sub-module. For convenience and readability, we denote a single training example as an image $I$ and a target tag sequence $Y$. The target tag sequence $Y$ contains training annotations $y_1,\dots,y_{T-1}$ and a special \textit{STOP} signal ${y_T}$. Similarly, an input tag sequence $X$ contains the \textit{START} signal $x_1$ followed by the training annotation tags $x_2,\dots,x_T$.

As shown in Figure \ref{fig_model}, we use RNN as a decoder that decodes tag sequence $Y$ from the input image $I$. To fit the image and tag sequence into the RNN model, we first embed them into latent image space and tag space with the image embedding and tag embedding submodule, respectively. Then we train our model with the embedded image and tag vectors. After training, the model will be able to generate a sequence of tags only from the (unseen) input image. 

\subsection{Image Embedding}
We either use pre-trained CNN features or jointly train a CNN to extract image features. In both cases we add a linear projection layer to project the output of CNN into $H$ dimensional space, where $H$ is the number of nodes in RNN hidden layer. In this way the CNN can be directly joined with the RNN sub-module. 

\subsection{Tag Embedding}
Instead of directly using one-hot vectors to represent tags, we map the tags to $D$ dimensional embedding vectors by using a lookup table like the common way to learn distributed word embeddings~\cite{mikolov2013efficient}. The lookup table is trainable and can learn what kind of representation to generate through training. In this way, the learned $D$ dimensional tag representation can be optimized for minimizing the annotation error. 

\subsection{Tag Sequence Generation}
We describe the tag sequence generation in two phases: training and testing.

In the training phase, the LSTM accepts an image embedding vector as its initial hidden state $h_0$ and the cell state $c_0$ of LSTM is initialized as zero. The \textit{START} signal is fed to the LSTM as its first input $x_1$. From the time step $t = 1$, the model will continue computing output score $s^t$ conditioned on $h_t$, then predicted tag index $\hat{y_t}$ will be decided by:
\begin{equation}
\label{tag classifier}
	\hat{y_t} =  \argmax_{j} s^t_j \;\;\; \mbox{for}\;\;\; j=1, \ldots, V
\end{equation}%
where $s^t_j$ is the score for tag index $j$ at time step $t$ and $V$ is the vocabulary size plus one (for $STOP$ signal).
On the other hand, $h_t$ is based on the current input $x_t$, the previous hidden state $h_{t-1}$ and cell state $c_{t-1}$. In this way, when predicting tags, the model can refer to both the current input tag and the previous predicted tags. The procedure that how hidden state and cell state propagate through time step is described as below:
{\setlength{\arraycolsep}{0.14em}
\begin{eqnarray}
	f_t &=& \sigma(W_f \cdot [h_{t-1}, x_t] + b_f)\\
	i_t &=& \sigma(W_i \cdot [h_{t-1}, x_t] + b_i)\\
	o_t &=& \sigma(W_o \cdot [h_{t-1}, x_t] + b_o)\\
	g_t &=& \tanh(W_g \cdot [h_t-1, x_t] + b_g)\\
	c_t &=& f_t \odot c_{t-1} + i_t \odot g_t\\
	h_t &=& o_t \odot \tanh(c_t)
\end{eqnarray}%
\setlength{\arraycolsep}{5pt}}%
where $f_t$, $i_t$, $o_t$ $g_t$ are the gate units of LSTM~\cite{hochreiter1997long}, and $W_*$, $b_*$ represent the corresponding weights and bias. The $\cdot$ and $\odot$ stand for the operator of matrix multiplication and element-wise multiplication respectively. The loss function of RIA is defined as the cross-entropy of prediction score $s^t$:
\begin{equation}
\label{loss}
	L =  \sum_{t=1}^{T}{-\log\frac{\exp{(s^t_{y_t})}}{\sum_{j=1}^{V}{\exp{(s^t_j)}}}}
\end{equation}

In the testing phase, referring to Figure \ref{fig_model}, the procedure is similar but simpler. The only needed input is the test image and the \textit{START} signal for triggering the first output tag. Then the sequence generation loop starts, in which the output of each time step $t$ will be used as the input of next time step $t+1$, until the \textit{STOP} signal is predicted.

\begin{table}[!t]
\caption{Dataset Description}
\label{dataset}
\centering
\begin{tabular}{l *{3}{c}}
	\toprule
	                 & Corel 5K    & ESP Game     & IAPR TC12\\
	\cmidrule{1-4}
	Vocabulary size  & 260         & 269          & 291\\
	Number of images & 4,493       &18,689        & 17,665\\
	Words per image  & 3.4 / 5     & 4.7 / 15     & 5.7 / 23\\
	Images per word   & 58.6 / 1004 & 362.7 / 4553 & 347.7 / 4999\\ 
	\bottomrule
\end{tabular}
\end{table}

\begin{table*}[!t]
\caption{Experimental Results of Arbitrary Length Annotation}
\label{arbitrary length}
\centering
\begin{ThreePartTable}
  \begin{tabu} to 140mm {X[6.1,l] X[2.5,c] *{3}{X[c] X[c] X[c] X[1.3,c]}}  
    \toprule
	              &          & \multicolumn{4}{c}{Corel 5K} & \multicolumn{4}{c}{ESP GAME} & \multicolumn{4}{c}{IAPR TC12}\\
	\cmidrule{3-14}
	Method        & Features & P  & R  & F  & N+           & P  & R  & F  & N+            & P  & R  & F  & N+\\
	\midrule
	RIA (dictionary) & fc7      & 30 & 29 & 30 & 138          & 32 & 29 & 29 & 249           & 32 & 28 & 29 & 239\\
	RIA (random)  & fc7      & {\bf34} & 34 & 32 & 139          & {\bf36} & 24 & 27 & 230           & 33 & 25 & 28 & 241\\
	RIA (rare-first)  & fc7      & 32 & {\bf35} & {\bf32} & {\bf139}          & 33 & 31 & {\bf31} & 249           & {\bf35} & 34 & {\bf34} & {\bf267}\\
	RIA (frequent-first) & fc7   & 30 & 30 & 29 & 126         & 34 & 23 & 24 & 216           & 31 & 20 & 22 & 207\\
	\midrule
	RIA (dictionary) & conv5    & 27 & 28 & 26 & 119          & 30 & 26 & 26 & 234           & 30 & 25 & 26 & 240\\
	RIA (random)  & conv5    & 28 & 29 & 27 & 127          & 29 & 22 & 25 & 233           & 30 & 20 & 23 & 222\\
	RIA (rare-first)  & conv5    & 32 & 33 & 30 & 134          & 31 & 28 & 29 & 243           & 32 & 29 & 30 & 258\\
	RIA (frequent-first) &conv5  & 28 & 29 & 27 & 125          & 30 & 22 & 24 & 218           & 29 & 19 & 21 & 200\\
	\midrule
	RIA (dictionary) & finetune & 26 & 29 & 26 & 128          & 31 & 30 & 29 & 251           & 32 & 34 & 31 & 261\\
	RIA (rare-first) & finetune & 31 & 33 & 31 & 135          & 33 & {\bf33} & 31 & {\bf251}           & 35 & {\bf37} & 34 & 265\\
	\bottomrule
\end{tabu}

\end{ThreePartTable}
\end{table*}

\begin{table*}[!t]
\caption{Experimental Results of Top-5 Annotation}
\label{top-5}
\centering
\begin{ThreePartTable}
  \begin{tabu} to 140mm {X[6.1,l] X[2.5,c] *{3}{X[c] X[c] X[c] X[1.3,c]}}  
    \toprule
	                 &          & \multicolumn{4}{c}{Corel 5K} & \multicolumn{4}{c}{ESP GAME} & \multicolumn{4}{c}{IAPR TC12}\\
	\cmidrule{3-14}
	Method           & Features & P  & R  & F  & N+           & P  & R  & F  & N+            & P  & R  & F  & N+\\
	\midrule
	MBRM\cite{feng2004multiple} & HC\tnote{1}  & 24 & 25 & 25 & 122   & 18 & 19 & 19 & 209   & 24 & 23 & 24 & 223\\
	JEC\cite{makadia2008new} & HC    & 27 & 32 & 29 & 139   & 22 & 25 & 23 & 224   & 28 & 29 & 29 & 250\\
	TagProp\cite{guillaumin2009tagprop} & HC & 33 & 42 & 37 &160    & 39 & 27 & 32 & 239   & 46 & 35 & 40 & 266\\
	2PKNN\cite{verma2012image}  & HC & 39 & 40 & 40 & 177   & 51 & 23 & 32 & 245   & 49 & 32 & 39 & 274\\
	 
	\midrule
	JEC           & fc7      & 31 & 32 & 31 & 141          & 26 & 22 & 24 & 234           & 28 & 21 & 24 & 237\\
	2PKNN         & fc7      & {\bf33}\tnote{2} & 30 & 32 & {\bf160}          & {\bf40} & 23 & 29 & {\bf250}           & {\bf38} & 23 & 29 & 261\\   
	\midrule
	RIA (dictionary) & fc7   & 30 & 29 & 30 & 138          & 32 & 27 & 27 & 241           & 31 & 26 & 27 & 233\\
	RIA (rare-first)     & fc7   & 32 & {\bf35} & {\bf32} & 139          & 32 & {\bf32} & {\bf31} & 249           & 35 & {\bf34} & \bf{33} & {\bf267}\\
	\bottomrule
  \end{tabu}
  	\vspace{0.7mm}	
	\begin{minipage}{137mm}
  	\begin{tablenotes}[flushleft]
		\footnotesize
		\item[1] \hspace{0.5mm} HC: hand-crafted features.
		\item[2] \hspace{0.5mm} For a fair comparison, we only use bold fonts for the highest value among the methods using the same fc7 features.
	\end{tablenotes}
	\end{minipage}

\end{ThreePartTable}
\end{table*}

\begin{figure*}[!t]
\centering
\subfloat[Corel 5K]{\includegraphics[width=.9\textwidth]{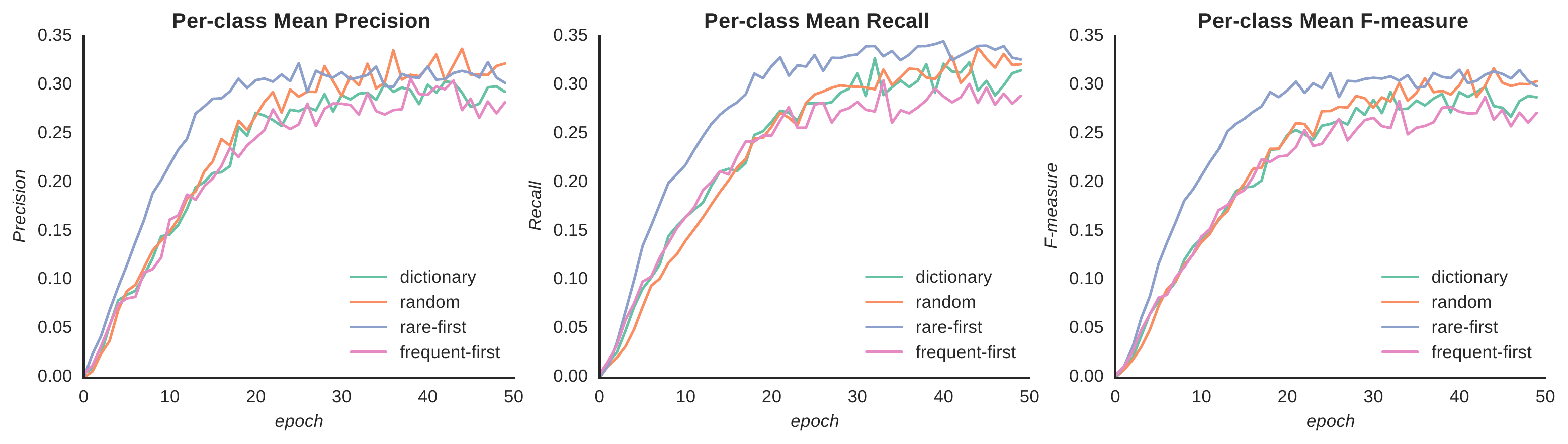}%
\label{corel5k}}
\hfil
\subfloat[ESP Game]{\includegraphics[width=.9\textwidth]{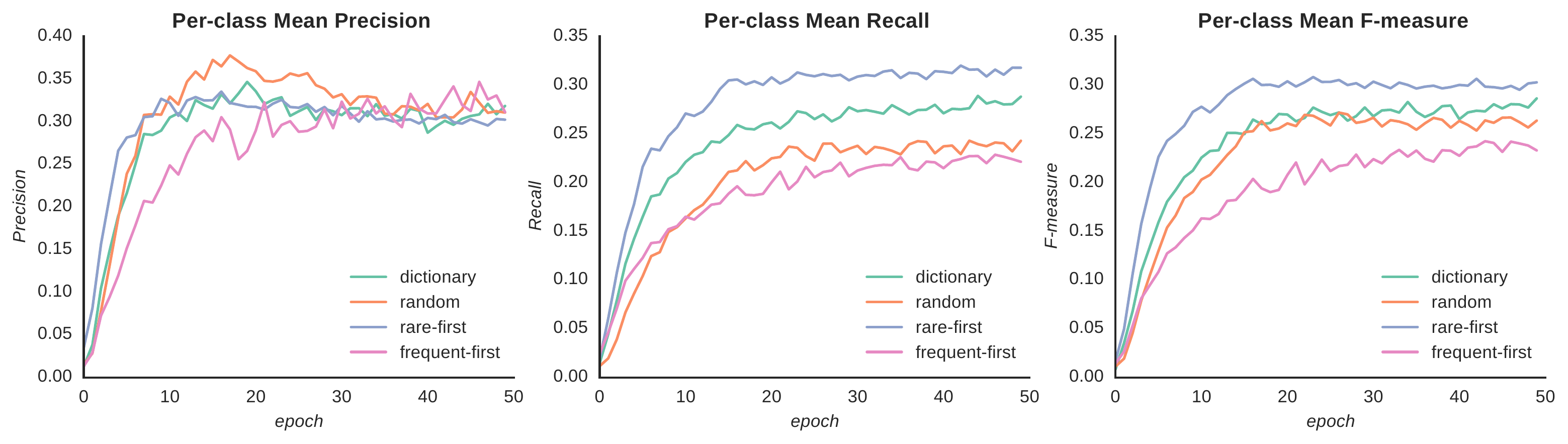}%
\label{espgame}}
\hfil
\subfloat[IAPR TC12]{\includegraphics[width=.9\textwidth]{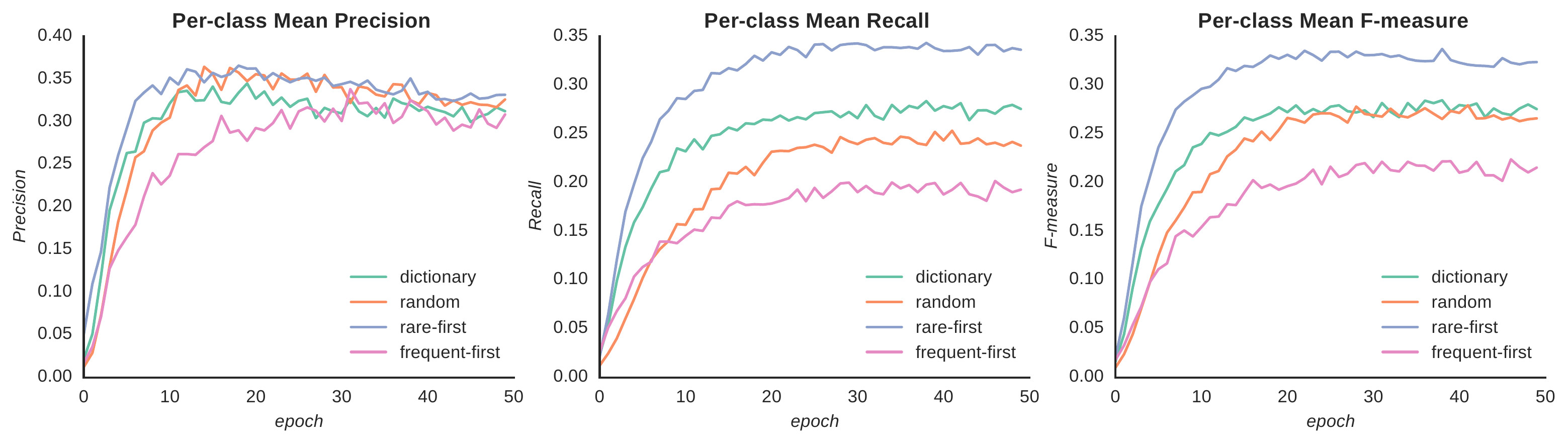}%
\label{iaprtc12}}
\caption{Arbitrary length annotation results on all datasets. We show the trained models of first 50 epochs for evaluation and comparison of different tag orders.}
\label{fig_result}
\end{figure*}

\subsection{Order of Tag Sequence}
To use the original training annotations as the input of LSTM, we have to sort the tag set to a tag sequence. We provide four orders: dictionary order, random order, rare-first order and frequent-first order. The dictionary order sorts the tags for each image alphabetically; the random order generates random tag sequence for each image as its name suggests; the rare-first order put the rarer tag before the more frequent ones (based on tag frequency in the dataset); the frequent-first order put the more frequent tag before the less frequent ones.

\section{Datasets and Experimental Setup}
In this section we first present the dataset used in our experiments, then we describe the different experimental settings and the evaluation measures for the experiments. Finally we explain the training details in our experiments.

\subsection{Datasets}
We adopt three image annotation datasets that have been used in previous work: Corel~5K~\cite{duygulu2002object}, ESP~Game~\cite{von2004labeling}, and IAPR~TC12~\cite{grubinger2007analysis}. Table \ref{dataset} shows statistics of the training sets of three datasets, some of which are described in a mean / maximum manner.

\subsection{Experimental Setting}
First, we compare RIA model with different tag sequence orders in the task of arbitrary length annotation. To further explore the image embedding submodule, we also compare the RIA models trained with different kinds of CNN features.

Second, we compare RIA model with existing methods on the three datasets in top-$5$ evaluation measures. For a fair comparison, especially we want to compare with the state-of-the-art methods that use the same CNN features as we adopt in our model. 

\subsection{Evaluation Measures}
For both top-$5$ annotation and arbitrary length annotation, we use precision $P$, recall $R$ and F-measure $F$ averaged over classes as the main evaluation measures. Another widely used measure $N+$, which represents the number of classes with non-zero recall value, is also reported. 

\subsection{Training Details}
We use three different ways to obtain the visual features: the last fully-connected layer of a pre-trained CNN denoted as \textit{fc7}, the last convolutional layer of a pre-trained CNN denoted as \textit{conv5}, and the output of a jointly trained (fine-tuned) CNN. The specific CNN model used here is the VGG-16 net~\cite{simonyan2014very}. For the tag sequence prediction module (LSTM), we set the dimension of hidden states $H$ and input $D$ both to be 1024, and we finally choose the number of hidden layers to be $1$ after exhaustive validations.

The learning rate policy used in our experiment is Adam~\cite{kingma2014adam}, which has been widely used recently. We set the initial learning rate, $\beta_1$, $\beta_2$ and $\epsilon$ as $0.0001$, $0.9$, $0.999$, $0.1$, respectively. Dropout with a ratio of $0.5$ is used in the tag classification layers of RNN. All the hyper-parameters are selected by cross-validation.  


\section{Experimental Results}
\subsection{Arbitrary Length Annotation}
Table \ref{arbitrary length} shows that \textit{fc7} features achieve better performance than \textit{conv5} features in our model. The fined-tuned CNN features have a similar performance to \textit{fc7} features, but need much more training time. Thus in the following experiments we only compare our models using \textit{fc7} features with other methods. Also, the rare-first order outperforms other orders in almost all evaluation measures. From Figure \ref{fig_result}, we observe that models using rare-first order converge faster than others, and the difference is even more significant in the larger datasets ESP~Game and IAPR~TC12. The random order has a slight advantage over other orders in precision, while in terms of recall it has very poor performance. For F-measure, dictionary and random order have similar performance. The frequent-first order has the worst performance in recall and F-measure.

We compare the experimental results with our expectation: First, though dictionary order actually assigns all the tags of training examples in the same rule, it is almost meaningless since it does not provide any semantic information about the images or tags, and thus it leads to a poor performance. Second, though random order provides some possible proper orders for each training example, it does not follow the same rule and makes the model confused about the noisy orders, which may also result in a low recall rate. Third, rare-first order considers the frequency of tags and to some extent can help handle the rare tags problem, which is very important for improving the per-class measures. Besides, it uses the same rule to sort tags of all training examples, hence makes it easy for the model to learn. Finally, the frequent-first order has worse performance than we expected. We analyze the reasons why frequent-first order performs poorly especially in large datasets: the frequent tags are usually easier to predict than rare tags, and the frequent-first order puts the frequent tags first, so easy work becomes easier, but hard work becomes more difficult, which causes the extremely low per-class mean recall rate. The lowest N+ score also indicates that frequent-first order harms the ability of the model to correctly predict rare tags.

Our experiments show that the order of tag sequence is crucial for tag sequence generation. However, note that we are only using several naive approaches to decide the order, and we believe that there should be better ways to choose or learn an optimal order for this task.

\subsection{Top-5 annotation}
As shown in Table \ref{top-5}, in the conventional top-5 annotation task, our model outperforms several state-of-the-art methods that use the same CNN features. Although the same methods with multiple hand-crafted features and metric learning have better performance, the advantage of using deep features is that we can avoid the complexity of hand-crafted features and the expensiveness of metric learning. Besides the comparable performance to several state-of-the-art methods, our model also runs in an extremely fast testing speed: 5~ms per image on an NVIDIA Titan X GPU. This is very difficult for KNN based methods to achieve, especially in large scale practical problems. That is because the testing time of KNN based methods is increasing linearly with the size of training examples, while the testing time of our model is constant, i.e., not affected by the dataset size.

\section{Conclusion}
We transformed the image annotation task to a sequence generation problem, and proposed a novel Recurrent Image Annotator model that receives an image as input and predicts a sequence of tags recursively. We evaluated our model in the traditional top-5 evaluation setting on three different image annotation datasets. The experimental results show that our model can achieve comparable performance to some state-of-the-art methods. On the condition of only using deep features without expensive metric learning, our model outperforms several state-of-the-art methods. We also evaluated our model on the arbitrary length annotation task, where the model has to decide appropriate annotation length automatically. To explore the influence of the tag sequence order used in the training phase, we evaluated several order candidates and our experiments confirmed the importance of a proper order in the tag sequence generation problem. From the empirical experimental results, we conclude that RNN model is capable of doing image annotation task, and since this is only a start for adopting RNN or other sequence generation techniques in this field, we believe that there is much more to explore in the future work.


\ifCLASSOPTIONcompsoc
\else
\fi




%
\bibliographystyle{IEEEtran}
\bibliography{IEEEabrv,recurrent_image_annotation}

\begin{thebibliography}{10}
\providecommand{\url}[1]{#1}
\csname url@samestyle\endcsname
\providecommand{\newblock}{\relax}
\providecommand{\bibinfo}[2]{#2}
\providecommand{\BIBentrySTDinterwordspacing}{\spaceskip=0pt\relax}
\providecommand{\BIBentryALTinterwordstretchfactor}{4}
\providecommand{\BIBentryALTinterwordspacing}{\spaceskip=\fontdimen2\font plus
\BIBentryALTinterwordstretchfactor\fontdimen3\font minus
  \fontdimen4\font\relax}
\providecommand{\BIBforeignlanguage}[2]{{%
\expandafter\ifx\csname l@#1\endcsname\relax
\typeout{** WARNING: IEEEtran.bst: No hyphenation pattern has been}%
\typeout{** loaded for the language `#1'. Using the pattern for}%
\typeout{** the default language instead.}%
\else
\language=\csname l@#1\endcsname
\fi
#2}}
\providecommand{\BIBdecl}{\relax}
\BIBdecl

\bibitem{zhang2012review}
D.~Zhang \emph{et~al.}, ``A review on automatic image annotation techniques,''
  \emph{Pattern Recognition}, vol.~45, no.~1, pp. 346--362, 2012.

\bibitem{feng2004multiple}
S.~Feng \emph{et~al.}, ``Multiple bernoulli relevance models for image and
  video annotation,'' in \emph{Proc. of IEEE CVPR}, vol.~2, 2004, pp.
  1002--1009.

\bibitem{makadia2008new}
A.~Makadia \emph{et~al.}, ``A new baseline for image annotation,'' in
  \emph{Computer Vision--ECCV 2008}.\hskip 1em plus 0.5em minus 0.4em\relax
  Springer, 2008, pp. 316--329.

\bibitem{guillaumin2009tagprop}
M.~Guillaumin \emph{et~al.}, ``Tagprop: Discriminative metric learning in
  nearest neighbor models for image auto-annotation,'' in \emph{Proc. of IEEE
  CVPR}, 2009, pp. 309--316.

\bibitem{verma2012image}
Y.~Verma and C.~Jawahar, ``Image annotation using metric learning in semantic
  neighbourhoods,'' in \emph{Computer Vision--ECCV 2012}.\hskip 1em plus 0.5em
  minus 0.4em\relax Springer, 2012, pp. 836--849.

\bibitem{bahdanau2014neural}
D.~Bahdanau \emph{et~al.}, ``Neural machine translation by jointly learning to
  align and translate,'' \emph{arXiv preprint arXiv:1409.0473}, 2014.

\bibitem{mao2014explain}
J.~Mao \emph{et~al.}, ``Explain images with multimodal recurrent neural
  networks,'' \emph{arXiv preprint arXiv:1410.1090}, 2014.

\bibitem{vinyals2015show}
O.~Vinyals \emph{et~al.}, ``Show and tell: A neural image caption generator,''
  in \emph{Proc. of IEEE CVPR}, 2015, pp. 3156--3164.

\bibitem{wang2016cnnrnn}
J.~Wang \emph{et~al.}, ``Cnn-rnn: A unified framework for multi-label image
  classification,'' \emph{axXiv preprint arXiv:1604.04573}, 2016.

\bibitem{duygulu2002object}
P.~Duygulu \emph{et~al.}, ``Object recognition as machine translation: Learning
  a lexicon for a fixed image vocabulary,'' in \emph{Computer Vision―ECCV
  2002}.\hskip 1em plus 0.5em minus 0.4em\relax Springer, 2002, pp. 97--112.

\bibitem{carneiro2007supervised}
G.~Carneiro \emph{et~al.}, ``Supervised learning of semantic classes for image
  annotation and retrieval,'' \emph{IEEE TPAMI}, vol.~29, no.~3, pp. 394--410,
  2007.

\bibitem{lavrenko2003model}
V.~Lavrenko \emph{et~al.}, ``A model for learning the semantics of pictures,''
  in \emph{Proc. of NIPS}, 2003, pp. 553--560.

\bibitem{cusano2003image}
C.~Cusano \emph{et~al.}, ``Image annotation using svm,'' in \emph{Electronic
  Imaging 2004}.\hskip 1em plus 0.5em minus 0.4em\relax International Society
  for Optics and Photonics, 2003, pp. 330--338.

\bibitem{grangier2008discriminative}
D.~Grangier and S.~Bengio, ``A discriminative kernel-based approach to rank
  images from text queries,'' \emph{IEEE TPAMI}, vol.~30, no.~8, pp.
  1371--1384, 2008.

\bibitem{gong2013deep}
Y.~Gong \emph{et~al.}, ``Deep convolutional ranking for multilabel image
  annotation,'' \emph{arXiv preprint arXiv:1312.4894}, 2013.

\bibitem{wei2014cnn}
Y.~Wei \emph{et~al.}, ``Cnn: Single-label to multi-label,'' \emph{arXiv
  preprint arXiv:1406.5726}, 2014.

\bibitem{lecun1998gradient}
Y.~LeCun \emph{et~al.}, ``Gradient-based learning applied to document
  recognition,'' \emph{Proceedings of the IEEE}, vol.~86, no.~11, pp.
  2278--2324, 1998.

\bibitem{krizhevsky2012imagenet}
A.~Krizhevsky \emph{et~al.}, ``Imagenet classification with deep convolutional
  neural networks,'' in \emph{Proc. of NIPS}, 2012, pp. 1097--1105.

\bibitem{simonyan2014very}
K.~Simonyan and A.~Zisserman, ``Very deep convolutional networks for
  large-scale image recognition,'' \emph{arXiv preprint arXiv:1409.1556}, 2014.

\bibitem{he2015deep}
K.~He \emph{et~al.}, ``Deep residual learning for image recognition,''
  \emph{arXiv preprint arXiv:1512.03385}, 2015.

\bibitem{murthy2015automatic}
V.~N. Murthy \emph{et~al.}, ``Automatic image annotation using deep learning
  representations,'' in \emph{Proc. of ACM ICMR}, 2015, pp. 603--606.

\bibitem{pascanu2012difficulty}
R.~Pascanu \emph{et~al.}, ``On the difficulty of training recurrent neural
  networks,'' \emph{arXiv preprint arXiv:1211.5063}, 2012.

\bibitem{hochreiter1997long}
S.~Hochreiter and J.~Schmidhuber, ``Long short-term memory,'' \emph{Neural
  computation}, vol.~9, no.~8, pp. 1735--1780, 1997.

\bibitem{cho2014learning}
K.~Cho \emph{et~al.}, ``Learning phrase representations using rnn
  encoder-decoder for statistical machine translation,'' \emph{arXiv preprint
  arXiv:1406.1078}, 2014.

\bibitem{chung2014empirical}
J.~Chung \emph{et~al.}, ``Empirical evaluation of gated recurrent neural
  networks on sequence modeling,'' \emph{arXiv preprint arXiv:1412.3555}, 2014.

\bibitem{mikolov2013efficient}
T.~Mikolov \emph{et~al.}, ``Efficient estimation of word representations in
  vector space,'' \emph{arXiv preprint arXiv:1301.3781}, 2013.

\bibitem{von2004labeling}
L.~Von~Ahn and L.~Dabbish, ``Labeling images with a computer game,'' in
  \emph{Proc. of ACM SIGCHI}, 2004, pp. 319--326.

\bibitem{grubinger2007analysis}
M.~Grubinger, ``Analysis and evaluation of visual information systems
  performance,'' Ph.D. dissertation, Victoria University, 2007.

\bibitem{kingma2014adam}
D.~Kingma and J.~Ba, ``Adam: A method for stochastic optimization,''
  \emph{arXiv preprint arXiv:1412.6980}, 2014.

\end{thebibliography}

\end{document}